\documentclass[journal,article,submit,pdftex,moreauthors]{Definitions/mdpi} 
\usepackage{lineno}
\usepackage{xcolor}
\firstpage{1} 
\pubvolume{1}
\issuenum{1}
\articlenumber{0}
\pubyear{2024}
\copyrightyear{2024}
\datereceived{ } 
\daterevised{ } 
\dateaccepted{ } 
\datepublished{ } 
\hreflink{https://doi.org/} 

\Title{Sampling Audit Evidence Using a Naive Bayes Classifier}

\TitleCitation{Sampling Audit Evidence Using a Naive Bayes Classifier}

\Author{Guang-Yih Sheu $^{1,\dagger,\ddagger}$\orcidA{0000-0003-1248-6207} and Nai-Ru Liu $^{2,\ddagger}$}

\AuthorNames{Guang Yih Sheu and Nai Ru Liu}

\AuthorCitation{Sheu. G. Y.; Liu, N. R.}

\address{%
$^{1}$ \quad Department of Innovative Application and Management/Accounting and Information System, Chang-Jung Christian University, Tainan, Taiwan; xsheu@hotmail.com\\
$^{2}$ \quad Department of Accounting and Information System, Chang-Jung Christian University, Tainan, Taiwan; 110b17727@mailst.cjcu.edu.tw}

\corres{Correspondence: xsheu@hotmail.com}

\firstnote{These authors contributed equally to this work.}
\linenumbers

\abstract{Taiwan's auditors have suffered from processing excessive audit data, including drawing audit evidence. This study advances sampling techniques by integrating machine learning with sampling. This machine learning integration helps avoid sampling bias, keep randomness and variability, and target risker samples. We first classify data using a Naive Bayes classifier into some classes. Next, a user-based, item-based, or hybrid approach is employed to draw audit evidence. The representativeness index is the primary metric for measuring its representativeness. The user-based approach samples data symmetric around the median of a class as audit evidence. It may be equivalent to a combination of monetary and variable samplings. The item-based approach represents asymmetric sampling based on posterior probabilities for obtaining risky samples as audit evidence. It may be identical to a combination of non-statistical and monetary samplings. Auditors can hybridize those user-based and item-based approaches to balance representativeness and riskiness in selecting audit evidence. Three experiments show that sampling using machine learning integration has the benefits of drawing unbiased samples, handling complex patterns, correlations, and unstructured data, and improving efficiency in sampling big data. However, the limitations are the classification accuracy output by machine learning algorithms and the range of prior probabilities.}


\begin{document}
\nolinenumbers

\setcounter{section}{0} 
\section{Introduction}

Taiwan's auditors have recently suffered from processing excessive data, including drawing audit evidence. This audit evidence refers to the information to support auditors' findings or conclusions about those excessive data. Auditors desire assistance from emerging technologies such as machine learning algorithms or software robots in completing the sampling. The overload of sampling excessive data causes Taiwan's small to medium accounting firms to need more young auditors to help accountants. They even ask Taiwan's universities to provide excellent accounting students as potential employees.

This study develops a Naive Bayes classifier (e.g., [1]) as a sampling tool. It is employed to help auditors generate audit evidence from a massive volume of data. For example, enterprises employ enterprise resource planning or information management systems to manage accounting data. They output a colossal amount of data each day. For economic reasons, auditing all data is almost impossible. Auditors rely on sampling methods to generate audit evidence. It denotes that auditors audit less than 100 \% of data; nevertheless, the sampling risk will occur correspondingly. It implies the likelihood that auditors' conclusions based on samples may differ from the conclusion made from the entire data.

A previous study [2] suggested applying a classification algorithm to mitigate the sampling risk in choosing audit evidence. This published research constructed a neural network to classify data into some classes and generate audit evidence from each class. If classification results are accurate, the corresponding audit evidence is representative.

However, we may have intelligent demands in drawing audit evidence. For example, financial accounts accepting frequent transactions are risky in a money laundering problem. Criminals may own these financial accounts to receive black money. An auditor will be grateful for sampling such risky financial accounts as audit evidence. We select a Naive Bayes classifier to complete those intelligent demands of generating audit evidence since it provides the relationships between members in a class. Other alternative classification algorithms cannot provide similar relationships.

Many published studies (e.g., [3-5]) attempted to integrate machine learning with sampling; however, the research interest of most was not auditing. Their goal was to develop unique sampling methods for improving the performance of machine learning algorithms in solving specific problems (e.g., [3]). Some studies (e.g., [4]) suggested sampling with machine learning in auditing; moreover, only some researchers (e.g., [5]) have indeed implemented machine learning-based sampling in auditing.

This study starts acquiring audit evidence by appending some columns to data to store the classification results of a Naive Bayes classifier. It next classifies data into some classes. Referring to existing sampling methods, we next implement a user-based, item-based, or hybrid approach to draw audit evidence. The representativeness index [6] is the primary metric for measuring whether audit evidence is representative. The user-based approach draws samples symmetric around the median of a class. It may be equivalent to a combination of monetary and variable sampling methods [7]. The item-based approach denotes the asymmetric sampling based on posterior probabilities for detecting riskier samples. It may be equivalent to combining non-statistical and monetary sampling methods [7]. Auditors may hybridize these user- and item-based approaches to balance the representativeness and riskiness in selecting audit evidence.

The remainder of this study has five sections. Section 2 presents a review of relevant studies to this study. Section 3 shows an integration of a Naive Bayes classifier with sampling. Section 4 presents three experiments for testing the resulting works in Section 3. Section 5 discusses the experimental results. Based on the previous two sections, Section 6 lists this study's conclusion and concluding remarks.

\section{Literature review}
As stated earlier, only some studies have sampled data using a machine learning algorithm in auditing. This sparsity leads to harassment in searching for advice to implement this study.

If the purpose is to improve the efficiency of auditing, some published studies (e.g., [5]) integrated machine learning with sampling for detecting anomalies. For example, Chen et al. [5] selected the ID3, CART, and C4.5 algorithms to find anomalies in financial transactions. Their results indicated that a machine learning algorithm can simplify the audit of financial transactions by efficiently exploring their attributes.

Schreyer et al. [8,9] constructed an autoencoder neural network to sample journal entries in their two papers. They fed attributes of those journal entries into the resulting autoencoder. However, Schreyer et al. plotted figures to describe the representatives of samples.

Lee [10] built another autoencoder neural network to sample taxpayers. Unlike Schreyer et al. [8,9], Lee calculated the reconstruction error to quantify the representativeness of samples. This metric measures the difference between input data and outputs reconstructed using samples. Lower reconstruction errors indicate better representativeness of original taxpayers. Besides, Lee [10] used the Aprior algorithm to find those taxpayers who may be valuable to sample together. If one taxpayer breaks some laws, other taxpayers may also be fraudulent.

Chen et al. [11] applied the random forest classifier, XGBoost algorithm, quadratic discriminant analysis, and support vector machines model to sample attributes of Bitcoin daily transaction data. These attributes contain the property and network, trading and market, attention, and gold spot prices. The goal of this previous research is to predict Bitcoin daily prices. Chen et al. [11] found that machine learning algorithms predicted more accurately Bitcoin 5-minute interval prices than statistical methods did.

Different from the above-mentioned four studies, Zhang and Trubey [3] designed under-sampling and over-sampling methods to highlight rare events in a money laundering problem. Their goal was improving the performance of machine learning algorithms in modeling money laundering events. Zhang and Trubey [3] adopted the Bayes logistic regression, decision tree, random forest classifier, support vector machines model, and artificial neural network. 

In fields other than auditing, three examples are listed: Liberty et al. [12] defined a specialized regression problem to calculate the probability of sampling each record of a browse dataset. The goal was to sample a small set of records over which evaluating aggregate queries can be done both efficiently and accurately. Deriving their solution to the regression problem employs a simple regularized empirical risk minimization algorithm. Liberty et al. [12] concluded that machine learning integration improved both uniform and standard stratified sampling methods.

Hollingsworth et al. [13] derived generative machine learning models to improve the computational efficiency in sampling high-dimensional parameter spaces. Their results achieve orders of magnitude improvements in sampling efficiency compared to a brute-force search.

Artrith et al. [14] combined a genetic algorithm and a specialized machine-learning potential based on artificial neural networks to quicken the sampling of amorphous and disordered materials. They found that machine learning integration decreased the required calculations in sampling.

Other relevant studies discussed the benefits or challenges of integrating a machine learning algorithm with the audit of data. These studies only encourage or remind the current study to notice these benefits or challenges. For example, Huang et al. [15] suggested that a machine learning algorithm may serve as a 'Black Box' to help an auditor. However, auditors may need help in mastering a machine learning algorithm. Furthermore, auditors may have a wrong understanding of the performance of a machine learning algorithm. This misunderstanding causes auditors to believe we can always obtain accurate classification or clustering of data using a machine learning algorithm. Besides, it improves effectiveness and cost efficiency, analyzes massive data sets, and reduces time spent on tasks. Therefore, we should ensure the performance of a machine learning algorithm is sufficiently good before applying it to aid auditors' work.

\section{Naive Bayes classifier}

This study applies a Naive Bayes classifier (e.g., [1]) to select audit evidence since this classification algorithm provides posterior probabilities to implement the selection. A Naive Bayes classifies data according to posterior probabilities. We may employ posterior probabilities to relate different members of a class.

Suppose ${(\mathbf{X}_1,C_1),(\mathbf{X}_2,C_2)\ldots,(\mathbf{X}_N,C_N)}$ denote $N$ items of data where $C_i$ is the class variable, $\mathbf{X}_i = (X_{i1}, X_{i2}\ldots, X_{in})$, and $X_{ij}\;(j = 1, 2\ldots, n)$ is the $j$-th attribute of $\mathbf{X}_i$ and $n$ is the total number of attributes.

A Naive Bayes classifier is a supervised multi-class classification algorithm. As shown in Figure 1, developing a Naive Bayes classifier considers Bayes' theorem with conditional independence assumption between every pair of variables:
\begin{equation}
\Pr\left({{{C}_{i}}\vert{{\mathbf{X}}_{j}}}\right){=}\frac{\Pr\left({{{\mathbf{X}}_{j}}\vert{{C}_{i}}}\right)\Pr\left({{C}_{i}}\right)}{\Pr\left({{\mathbf{X}}_{j}}\right)}
\end{equation}
in which $i,j = 1, 2\ldots,N$, $\Pr\left({{{C}_{i}}\vert{{\mathbf{X}}_{j}}}\right)$ is the posterior probability, $\Pr\left({{{\mathbf{X}}_{j}}\vert{{C}_{i}}}\right)$ denotes the likelihood, $\Pr\left({{C}_{i}}\right)$ and $\Pr\left({{\mathbf{X}}_{j}}\right)$ is the prior probability.

Applying the assumption of features $X_{i1}, X_{i2}\ldots, X_{in}$ are independent of each other yields
\begin{equation}
\Pr\left({{{C}_{i}}\vert{{\mathbf{X}}_{j}}}\right){=}\frac{\Pr\left({{C}_{i}}\right)\mathop{\prod}\limits_{{k}{=}{1}}\limits^{n}{\Pr\left({{{\mathbf{X}}_{jk}}\vert{{C}_{i}}}\right)}}{\Pr\left({{\mathbf{X}}_{j}}\right)}
\end{equation}
where $i,j = 1, 2..., N$. Since the denominator of Equation (2) is the same for all $C_i$ classes, comparing the numerator of it for each $C_i$ class is implemented in classifying features $\mathbf{X}_1, \mathbf{X}_2,\ldots,\mathbf{X}_N$, $X_{ij}\;(j = 1, 2\ldots, n)$. This comparison ends when Equations. (3)-(4) are satisfied:
\begin{equation}
\Pr\left({{{C}_{i}}\vert{{\mathbf{X}}_{j}}}\right)\propto\Pr\left({{C}_{i}}\right)\mathop{\prod}\limits_{{k}{=}{1}}\limits^{n}{\Pr\left({{{\mathbf{X}}_{jk}}\vert{{C}_{i}}}\right)}
\end{equation}
\begin{equation}
\widehat{y}\in\mathop{\text{argmax}}\limits_{{i}\in\left\{{{1}{,}{2}\ldots{,}{N}}\right\}}\left[{\Pr\left({{C}_{i}}\right)\mathop{\prod}\limits_{{k}{=}{1}}\limits^{n}{\Pr\left({{{\mathbf{X}}_{jk}}\vert{{C}_{i}}}\right)}}\right]
\end{equation}
where $\widehat{y}$ denotes a class variable.

\begin{figure}[H]
\includegraphics[width=8 cm]{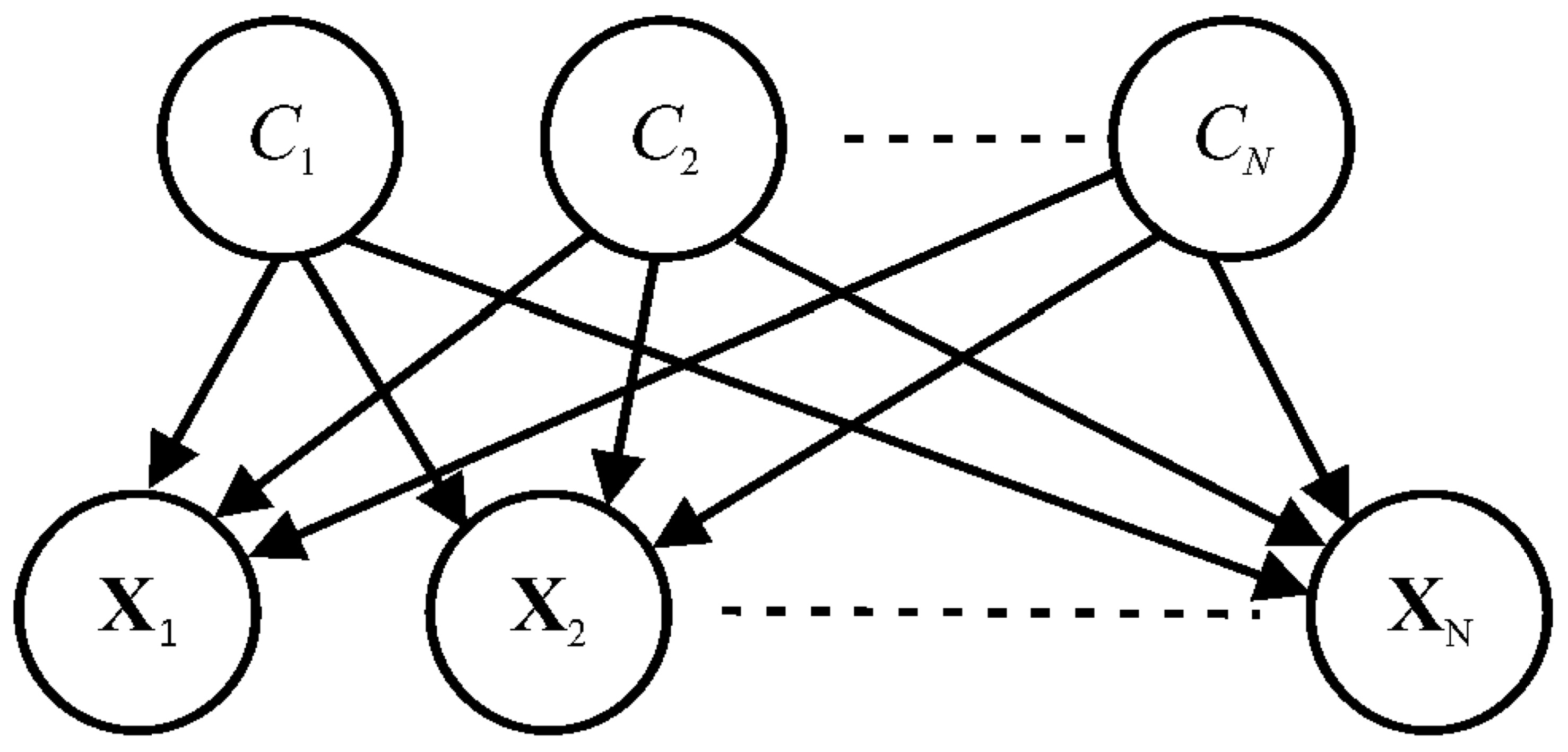}
\caption{Bayes' theorem}
\end{figure}
Regarding conventional sampling methods [7], this study designs user-based and item-based approaches in integrating Equations (3)-(4) with the selection of audit evidence:
\begin{enumerate}
\item [i.] User-based approach: In an attempt to generate unbiased representations of data, classifying ${(\mathbf{X}_1,C_1),(\mathbf{X}_2,C_2)\ldots,(\mathbf{X}_N,C_N)}$ and compute two percentile symmetric around the median of each class according to an auditor's professional preferences. Draw the $\mathbf{X}_1, \mathbf{X}_2,\ldots,\mathbf{X}_N$ bounded by the resulting percentiles as audit evidence, and
\item[ii.] Item-based approach: Suppose the ${\mathbf{X}_j,C_j}\;(1 \leq j \leq N)$ represent risky samples. Asymmetrically sample them based on the $\Pr\left({{{C}_{i}}\vert{{\mathbf{X}}_{j}}}\right)\;(1 \leq i \leq N)$ values as audit evidence after classifying ${(\mathbf{X}_1,C_1),(\mathbf{X}_2,C_2)\ldots,(\mathbf{X}_N,C_N)}$.
\end{enumerate}
\subsection{User-based approach}
Suppose the ${C}_{i}{\;(}{1}\leq{i}\leq{N}{)}$ is a class after classifying ${(\mathbf{X}_1,C_1),(\mathbf{X}_2,C_2)\ldots,(\mathbf{X}_N,C_N)}$, For implementing this classification, we compute posterior probabilities $\Pr\left({{{C}_{i}}\vert{{\mathbf{X}}_{j}}}\right)$ and regress the resulting $\Pr\left({{{C}_{i}}\vert{{\mathbf{X}}_{j}}}\right)$ values by a posterior probability distribution. Figure 2 shows an example. Deriving the detailed expression of this posterior probability distribution is unnecessary since deriving such an expression is not our goal. On the curve in Figure 2, we can determine two percentiles symmetric around the median. Draw samples $\mathbf{X}_L, \mathbf{X}_{L+1},\ldots,\mathbf{X}_M$ bounded by the resulting percentiles audit evidence. In mathematical formulations, the present user-based approach implements the following Equation (5) to output audit evidence:
\begin{equation}
{\mathbf{P}}_{{-}}\leq{\mathbf{X}}_{L}{,}{\mathbf{X}}_{{L}{+}{1}}\ldots{,}{\mathbf{X}}_{M}\leq{\mathbf{P}}_{{+}}
\end{equation}
where $\mathbf{P}_+$ and $\mathbf{P}_-$ are two percentiles defining this confidence interval.

Auditors may have unique preferences of percentiles $\mathbf{P}_+$ and $\mathbf{P}_-$. For example, if $\mathbf{P}_+$ and $\mathbf{P}_-$ are 97.5th and 2.5th percentiles, samples $\mathbf{X}_L, \mathbf{X}_{L+1},\ldots,\mathbf{X}_M$ represent audit evidence in a 95\% confidence interval.

Furthermore, computing posterior probabilities of samples $\mathbf{X}_L, \mathbf{X}_{L+1},\ldots,\mathbf{X}_M$ yields
\begin{equation}
\begin{array}{l}
{\Pr\left({{{C}_{i}}\vert{{\mathbf{P}}_{{-}}\leq{\mathbf{X}}_{L}{,}{\mathbf{X}}_{{L}{+}{1}}\ldots{,}{\mathbf{X}}_{M}}\leq{\mathbf{P}}_{{+}}}\right){=}}\\
{\Pr\left({{{C}_{i}}\vert{{\mathbf{X}}_{L}}}\right){+}\Pr\left({{{C}_{i}}\vert{{\mathbf{X}}_{{L}{+}{1}}}}\right){+}\ldots{+}\Pr\left({{{C}_{i}}\vert{{\mathbf{X}}_{M}}}\right){=}\mathop{\sum}\limits_{{k}{=}{L}}\limits^{M}{\Pr\left({{{C}_{i}}\vert{{\mathbf{X}}_{k}}}\right)}}
\end{array}
\end{equation}
\begin{figure}[H]
\includegraphics[width=9 cm]{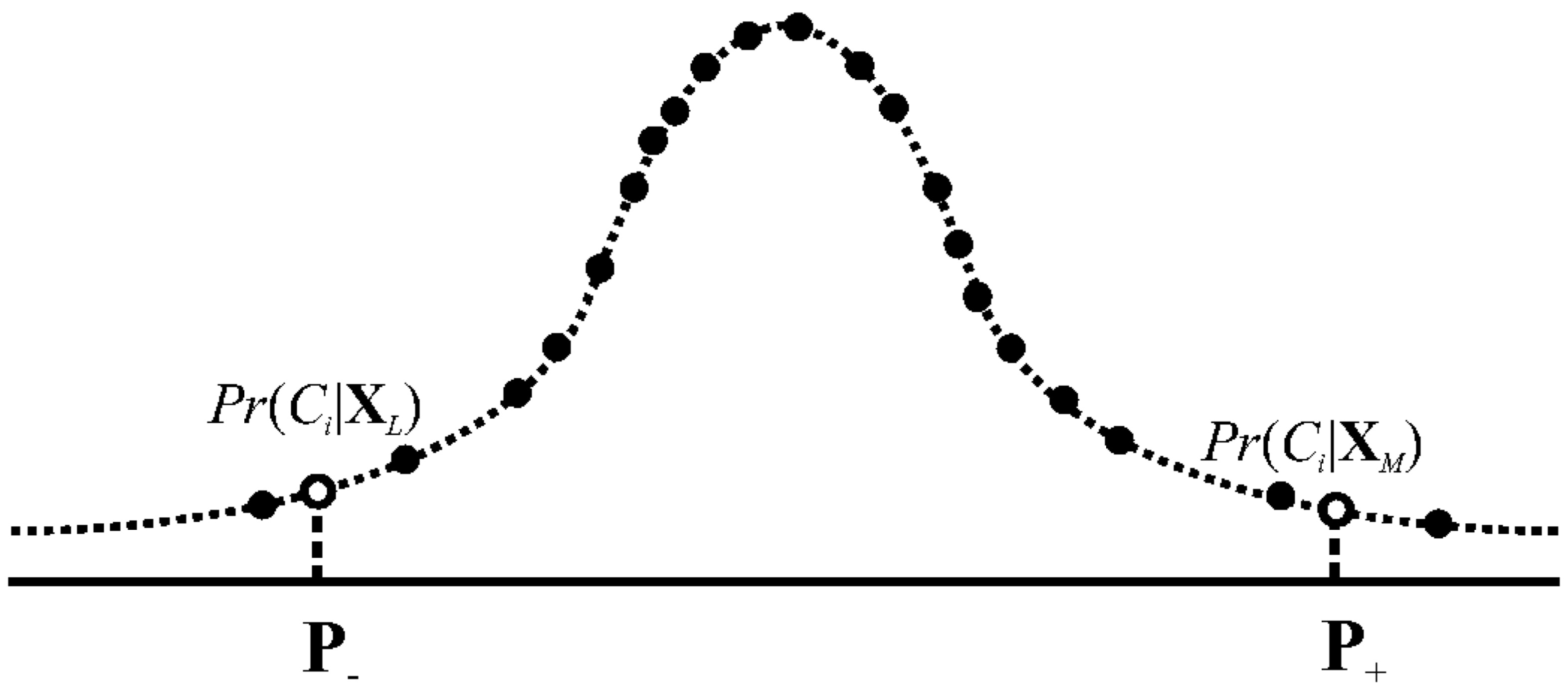}
\caption{Construction of a posterior probability distribution}
\end{figure}

After drawing audit evidence, this study measures the representativeness of these $\mathbf{X}_L, \mathbf{X}_{L+1},\ldots,\mathbf{X}_M$ by [3]
\begin{equation}
{\mathrm{Representativeness}}\hspace{0.33em}{\mathrm{index}}\hspace{0.33em}{(}{RI}{)}{=}{1}{-}\frac{12N({C}_{i})}{{4}{\left[{N({C}_{i})}\right]}^{2}{-}{1}}\mathop{\sum}\limits_{{r}{=}{L}}\limits^{M}{\left[{{F}\left({{X}_{L}}\right){-}\frac{{2}{r}{-}{1}}{2N({C}_{i})}}\right]}
\end{equation}
in which $i = 1, 2\ldots, N$, $N(C_i)$ is the total number of members in the $C_i$ class, and $F$ is the cumulative distribution function of the curve in Figure 2. Since $\mathbf{X}_L, \mathbf{X}_{L+1},\ldots,\mathbf{X}_M$ are discrete, this $F$ function is equal to
\begin{equation}
{F}\left({{X}_{r}}\right){=}\mathop{\sum}\limits_{{k}{=}{L}}\limits^{r}{\Pr\left({{{C}_{i}}\vert{{\mathbf{X}}_{k}}}\right)}
\end{equation}
where $L \leq i \leq M$. If total members in the $C_i\;(1 \leq I \leq N)$ class are sampled, the representativeness index $RI$ is identical to 1. On this $RI$ value, the goal of drawing audit evidence may be choosing sufficient samples but maintaining high $RI$ values.

Regarding existing audit sampling methods [4], the present user-based approach may be identical to a combination of the monetary and variable sampling methods.

\subsection{Item-based approach}
Similarly manipulating Sec. 3.1, suppose a $C_i\;(1 \leq i \leq N)$ is one of the classes resulting from the classification of data in which $\mathbf{X}_L, \mathbf{X}_{L+1},\ldots,\mathbf{X}_M$ are members of this $C_i$ class.

If we have a null hypothesis $H_0$ that members of the $C_i\;(1 \leq i \leq N)$ class are risky, a member $\mathbf{X}_L\;(1 \leq L \leq N)$ of this $C_i$ class with a lower $\Pr\left({{C}_{i}|{\mathbf{X}}_{k}}\right)$ value increases the possibility of rejecting this $H_0$. Hence, drawing this $\mathbf{X}_L$ as an audit evidence is valueless. To strengthen the belief that $H_0$ is true, it is better to sample asymmetrically members satisfying:
\begin{equation}
{0}{<}{\sigma}_{1}\leq\Pr\left({{C}_{i}|{\mathbf{X}}_{k}}\right){\leq}{1}
\end{equation}
where $L \leq k \leq M$ and $\sigma_1$ represents a selected threshold.

Furthermore, samples $\mathbf{X}_L$ and $\mathbf{X}_M$ may be simultaneously risky. Selecting them as audit evidence may be valuable. This selection may be based on the posterior probabilities of $\mathbf{X}_L\cap\mathbf{X}_M$:
\begin{equation}
\Pr\left({{C}_{i}{|}{\mathbf{X}}_{L}\cap{\mathbf{X}}_{M}}\right){=}\frac{\Pr\left({{\mathbf{X}}_{L}\cap{\mathbf{X}}_{M}{|}{C}_{i}}\right)\Pr\left({{C}_{i}}\right)}{\Pr\left({{\mathbf{X}}_{L}\cap{\mathbf{X}}_{M}}\right)}
\end{equation}

Further simplifying Equation (10) results in
\begin{equation}
\Pr\left({{{C}_{i}}\vert{{\mathbf{X}}_{L}\cap{\mathbf{X}}_{M}}}\right){=}\frac{\Pr\left({{\mathbf{X}}_{L}|{C}_{i}}\right)\Pr\left({{\mathbf{X}}_{M}|{C}_{i}}\right)\Pr\left({{C}_{i}}\right)}{\Pr\left({{\mathbf{X}}_{L}}\right)\Pr\left({{\mathbf{X}}_{M}}\right)}{=}\frac{\Pr\left({{C}_{i}|{\mathbf{X}}_{L}}\right)\Pr\left({{C}_{i}|{\mathbf{X}}_{M}}\right)}{\Pr\left({{C}_{i}}\right)}
\end{equation}

Samples satisfying ${0}{<}{\sigma}_{2}\leq\Pr\left({{{C}_{i}}\vert{{\mathbf{X}}_{L}\cap{\mathbf{X}}_{M}}}\right)\leq\frac{1}{\Pr\left({{C}_{i}}\right)}$ are drawn as audit evidence in which $\sigma_2$ is another selected threshold. The upper bound of Equation (11) depends upon the $Pr(C_i)$ value. To save time in searching those ${\mathbf{X}_L,\mathbf{X}_{M}}$ suitable for applying Equation (11), the Apriori algorithm states that we may start the search from those samples satisfying Equation (9). Such audit evidence may produce larger numerators in the last expression of Equation (11).

Furthermore, extending Equation (10) to samples $\mathbf{X}_L, \mathbf{X}_{L+1},\ldots,\mathbf{X}_M$ yields
\begin{equation}
\begin{array}{l}
{\Pr\left({{{C}_{i}}\vert{{\mathbf{X}}_{L}\cap{\mathbf{X}}_{L}\cap{...}\cap{\mathbf{X}}_{M}}}\right){=}}\\
{{=}\frac{\Pr\left({{{\mathbf{X}}_{L}}\vert{{C}_{i}}}\right)\Pr\left({{{\mathbf{X}}_{{L}{+}{1}}}\vert{{C}_{i}}}\right)\times{...}\times\Pr\left({{{\mathbf{X}}_{M}}\vert{{C}_{i}}}\right)\Pr\left({{C}_{i}}\right)}{\Pr\left({{\mathbf{X}}_{L}}\right)\Pr\left({{\mathbf{X}}_{{L}{+}{1}}}\right)\times{...}\times\Pr\left({{\mathbf{X}}_{M}}\right)}{=}\frac{\Pr\left({{{C}_{i}}\vert{{\mathbf{X}}_{L}}}\right)\Pr\left({{{C}_{i}}\vert{{\mathbf{X}}_{{L}{+}{1}}}}\right)\times{...}\times\Pr\left({{{C}_{i}}\vert{{\mathbf{X}}_{M}}}\right)}{{\left[{\Pr\left({{C}_{i}}\right)}\right]}^{{M}{-}{L}}}}
\end{array}
\end{equation}

Samples satisfying ${0}{<}{\mathit{\sigma}}_{3}\leq\Pr\left({{{C}_{i}}\vert{{\mathbf{X}}_{L}\cap{\mathbf{X}}_{L}\cap{...}\cap{\mathbf{X}}_{M}}}\right)\leq\frac{1}{{\left[{\Pr\left({{C}_{i}}\right)}\right]}^{{M}{-}{L}}}$ are selected as audit evidence in which $\sigma_3$ denotes third chosen threshold. Similarly, the upper bound of Equation (12) depends upon the ${\left[{\Pr\left({{C}_{i}}\right)}\right]}^{{L}{-}{M}}$ value. Again, the Apriori algorithm suggests that we can choose samples from those satisfying $\Pr\left({{{C}_{i}}\vert{{\mathbf{X}}_{L}\cap{\mathbf{X}}_{M}}}\right)\geq\sigma_2$.

Regarding existing audit sampling methods [4], the present item-based approach may be equivalent to a combination of non-statistical and monetary sampling methods.

Like Section 3.1, we calculate the representativeness index $RI$ [3] to check whether audit evidence is sufficiently representative.

\subsection{Hybrid approach}
Auditors may hybridize the resulting works in Sections 3.1-3.2 to balance representativeness and riskiness. We first apply the user-based approach to sample representative members bounded by two percentiles symmetric around the median of a $C_i\;(1\leq i \leq N)$ class. Applying the item-based approach to sample asymmetrically risker samples is next performed among those resulting representative samples.

\section{Results}

This study generates three experiments to illustrate the benefits and limitations of combining a machine learning algorithm with sampling. The first experiment demonstrates that machine learning integration helps avoid sampling bias and maintains randomness and variability. The second experiment shows that the proposed works help sample unstructured data. The final experiment shows that the hybrid approach balances representativeness and riskiness in sampling audit evidence.

Referring to the previous study [15], implementing machine learning integration with sampling is better based on the accurate classification results provided by a machine learning algorithm. Therefore, this study chooses a random forest classifier and a support vector machines model with a radial basis function kernel as baseline models.

\subsection{Experiment 1}

A customer ad click prediction data set contains $10^3$ (i.e., $N = 10^3$) records in which 50\% of customers clicked the advertisement and the remaining 50\% did not. This study uses the 'Daily time spent on site,' 'Age,' 'Area income,' 'Daily internet usage,' and 'Clicked on Ad' columns as experimental data. Two-thirds of those $10^3$ records are randomly chosen as train data, whereas others are test data. The 'Daily time spent on site,' 'Age,' 'Area income,' and 'Daily internet usage' columns are attributes $X_{ij}, (i = 1-4, j = 1, 2\ldots, N)$. Besides, set the class variable $C_j$ to indicate the 'Clicked on Ad' column equal to 'Clicked' or 'Not clicked'. Figure 3 shows variations of those $X_{ij}$ values.
\begin{figure}[H]
\includegraphics[width=8 cm]{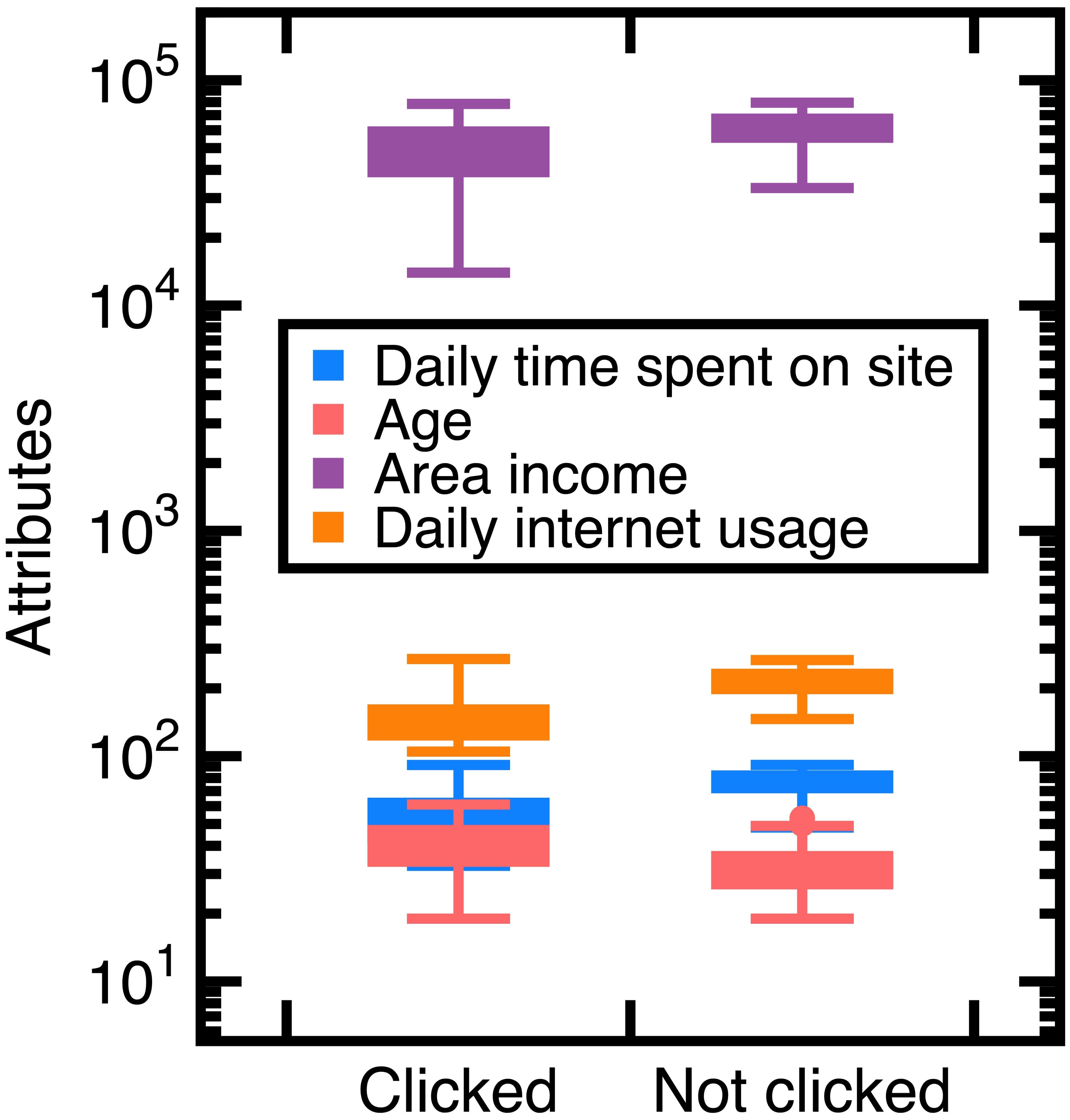}
\caption{Distributions of attributes $X_{ij}, (i = 1-4, j = 1, 2\ldots,N$) values in Experiment 1}
\end{figure} 
To avoid sampling frame errors [15], studying the classification accuracy output by Equations (3)-(4) is necessary. Figure 4 shows the resulting ROC curves in which NB, RF, and SVM are abbreviations of Naive Bayes, random forest, and support vector machines. This figure also shows the confusion matrix output by Equations (3)-(4). Its components have been normalized based on the amount of test data. Moreover, this study computes:
\begin{equation}
{\mathrm{accuracy}}{=}\frac{{\mathrm{true}}\;{\mathrm{positive}}{+}{\mathrm{true}}\;{\mathrm{negative}}}{{\mathrm{all}}\;{\mathrm{samples}}}{=}{0}{.}{964}
\end{equation}
\begin{equation}
{\mathrm{precision}}{=}\frac{{\mathrm{true}}\;{\mathrm{positive}}}{{\mathrm{true}}\;{\mathrm{positive}}{+}{\mathrm{false}}\;{\mathrm{positive}}}{=}{0}{.}{977}
\end{equation}
\begin{equation}
{\mathrm{recall}}{=}\frac{{\mathrm{true}}\;{\mathrm{positive}}}{{\mathrm{true}}\;{\mathrm{positive}}{+}{\mathrm{false}}\;{\mathrm{negative}}}{=}{0}{.}{956}
\end{equation}
\begin{equation}
{\mathrm{specificity}}{=}\frac{{\mathrm{true}}\;{\mathrm{negative}}}{{\mathrm{true}}\;{\mathrm{negative}}{+}{\mathrm{false}}\;{\mathrm{positive}}}{=}{0}{.}{974}
\end{equation}

Further computing the F1 score from Equations (14)-(15) yields
\begin{equation}
{\mathrm{F}}{1}\;{\mathrm{score}}{=}\frac{{2}\times{\mathrm{precision}}\times{\mathrm{recall}}}{{\mathrm{precision}}{+}{\mathrm{recall}}}{=}{0}{.}{965}
\end{equation}

Meanwhile, calculating the AUC from Figure 4 obtains 0.965 (Equations (3)-(4)), 0.953 (Random forest classifier), and 0.955 (Support vector machines model with a radial basis function kernel). These AUC values indicate that Equations (3)-(4) slightly outperform the random forest classifier and support vector machines model with a radial basis function kernel in avoiding sampling frame errors and undercoverage. However, all three algorithms are good models.
\begin{figure}[H]
\includegraphics[width=8 cm]{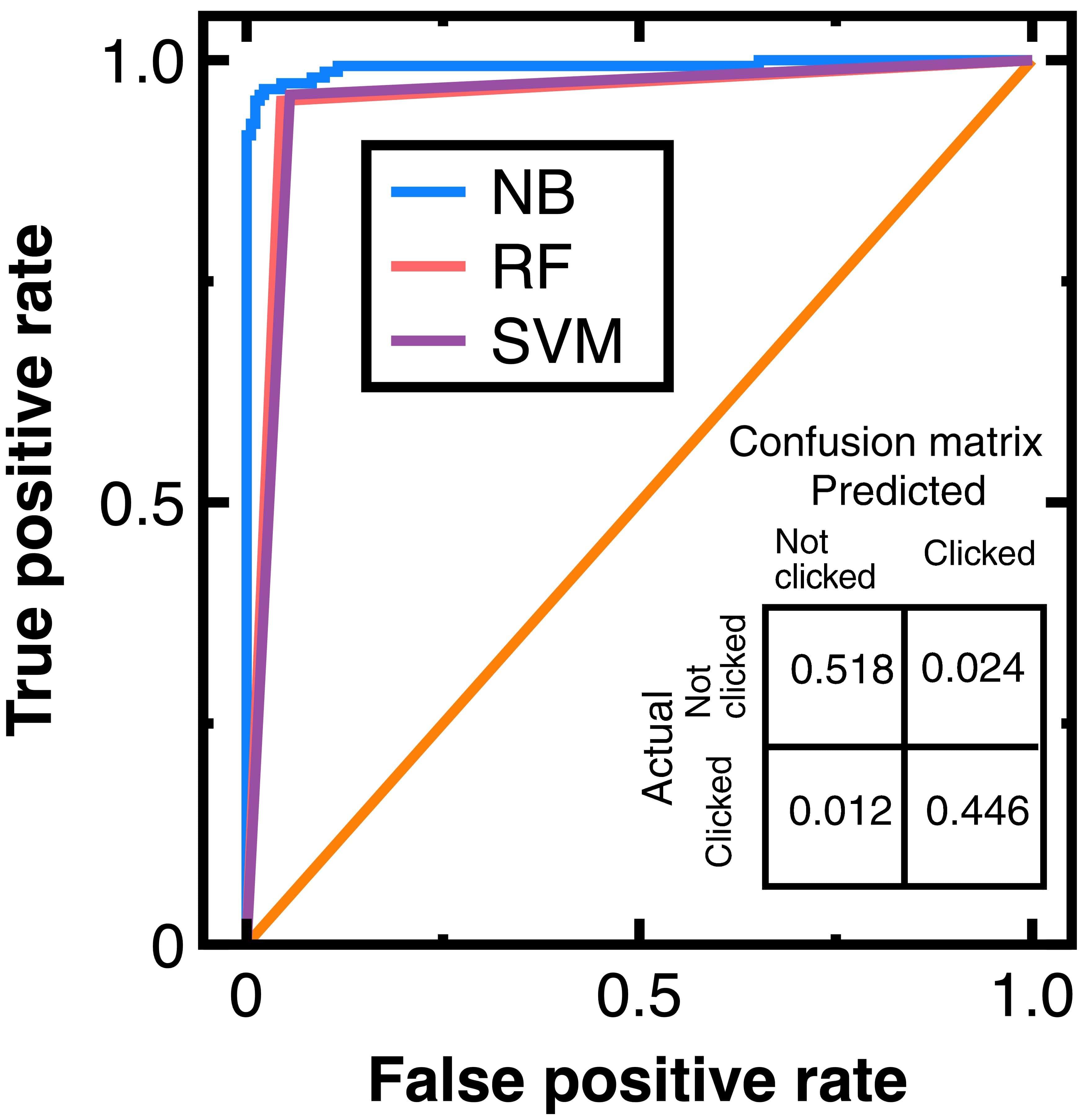}
\caption{ROC curves provided by different machine learning algorithms and the confusion matrix output by Equations (3)-(4) for Experiment 1}
\end{figure}

Our aim for testing Section 3.1 is to sample an unbiased representation of experimental data with machine learning integration. Figure 5 shows the resulting audit evidence with a 50 \% confidence interval for each class. Histograms on this figure's top and right sides compare the distributions of original customers and audit evidence. In this figure, light and heavy gray points denote experimental data, whereas red and blue colors mark audit evidence. The total number of blue and red points in Figure 5 equal 250, respectively. Substituting the resulting audit evidence into Equation (7) obtains the representativeness indices $RI$ listed in the legend of Figure 5.
\begin{figure}[H]
\includegraphics[width=8 cm]{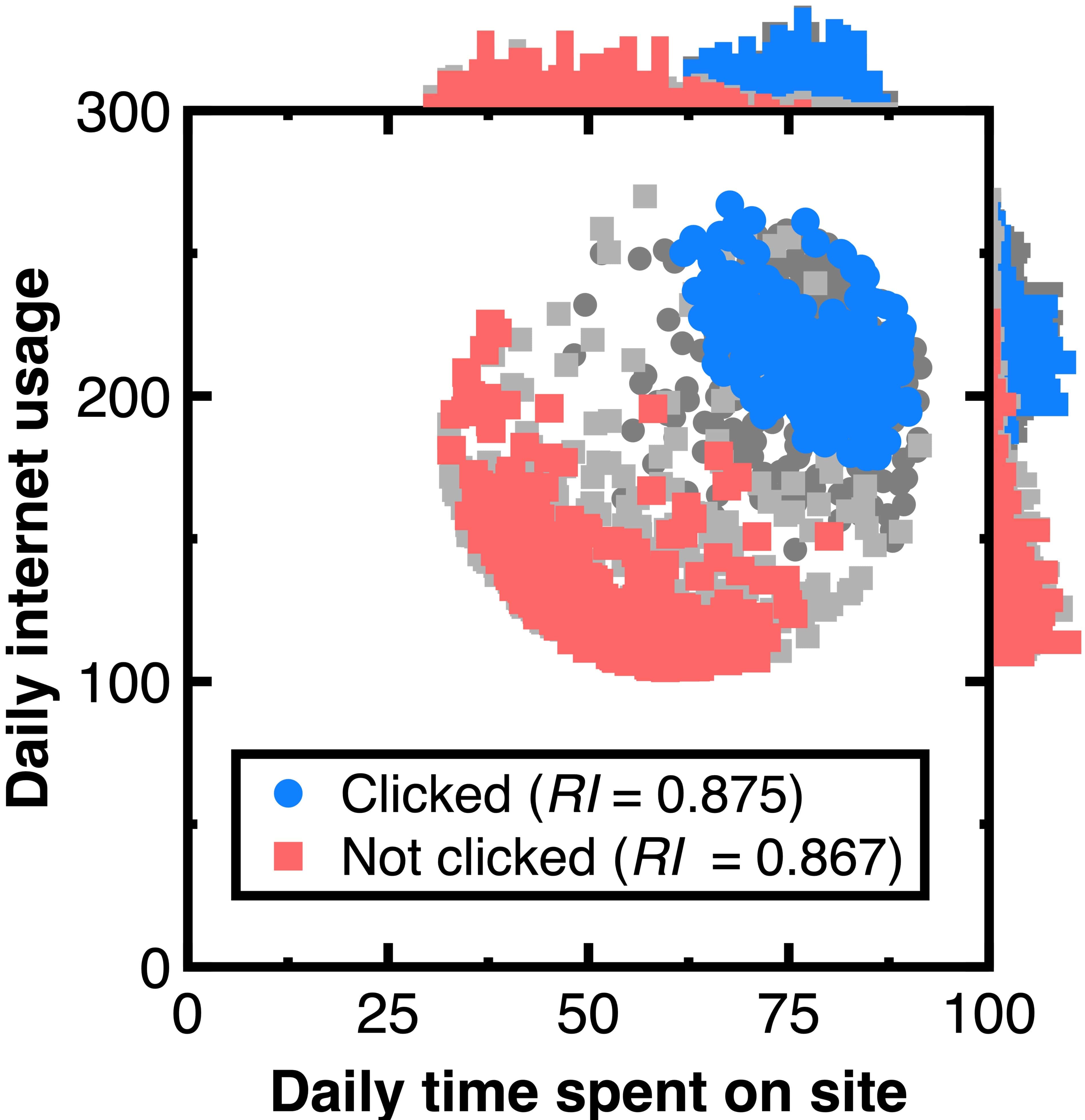}
\caption{Audit evidence for 50 \% confidence intervals}
\end{figure}

Suppose the null hypothesis defines that the experimental data and audit evidence originate from the same probability distribution. We calculate the Kolmogorov-Smirnov test statistic [16] to quantify the possibility of rejecting this null hypothesis. The result is equal to 0.044, and it is less than the critical value equal to $0.055\cong\frac{1.22}{\sqrt{500}}$ [16] for concluding Kolmogorov-Smirnov test statistics with considering the probability of 10\% in rejecting the null hypothesize. 

Calculating the Kolmogorov-Smirnov test statistic ensures that the audit evidence in Figure 5 is unbiased and representative of original customers. If the resulting Kolmogorov-Smirnov test statistic is lower than the critical value for concluding this test statistic, the original customers and audit evidence originate from the same probability distribution. Thus, we can reduce the risk of system errors or biases in estimating customers' attributes.

We have another aim of keeping the variability in testing Section 3.2. As marked by a blue cross in Figure 6, choose a customer with the predicted posterior probability of 0.999. The caption of Figure 6 lists the attributes of this customer. Other customers relevant to this customer are drawn as audit evidence and marked using red points in Figure 6. Besides, we still use light or heavy gray points representing the experimental data and histograms besides Figure 6 to describe the distribution of audit evidence. Since the denominator $Pr(C_i)$ of Equation (11) equals 0.5. setting the $\sigma_2$ threshold to 1.9999 is considered. Substituting the resulting audit evidence into Equation (7) yields the representativeness index $RI$ in the legend of Figure 7. Counting the number of drawn audit evidence yields 294.

Table 1 compares variability between the original 'Daily Internet use' variable and audit evidence. We employ the range, standard deviation, interquartile range, and coefficient of variation to measure the variability.

Measuring the variability helps understand the shape and spread of audit evidence. Table 1 shows that the audit evidence maintains the variability.
\begin{table}[H] 
	\caption{Comparison of the variability between original customers and audit evidence}
	\newcolumntype{C}{>{\centering\arraybackslash}X}
	\begin{tabularx}{\textwidth}{CCC}
		\toprule
			& \textbf{Original data}	& \textbf{Audit evidence}\\
		\midrule
		Range & [104.78,225.24] & [104.78,225.24]\\
		Standard deviation & 24.55 & 24.53 \\
		Interquartile range & 34.58 & 34.58 \\
		Skewness & 0.674 & 0.673 \\
		Coefficient of variation & 0.1731 & 0.173 \\
		\bottomrule
	\end{tabularx}
\end{table} 
\begin{figure}[H]
\includegraphics[width=8 cm]{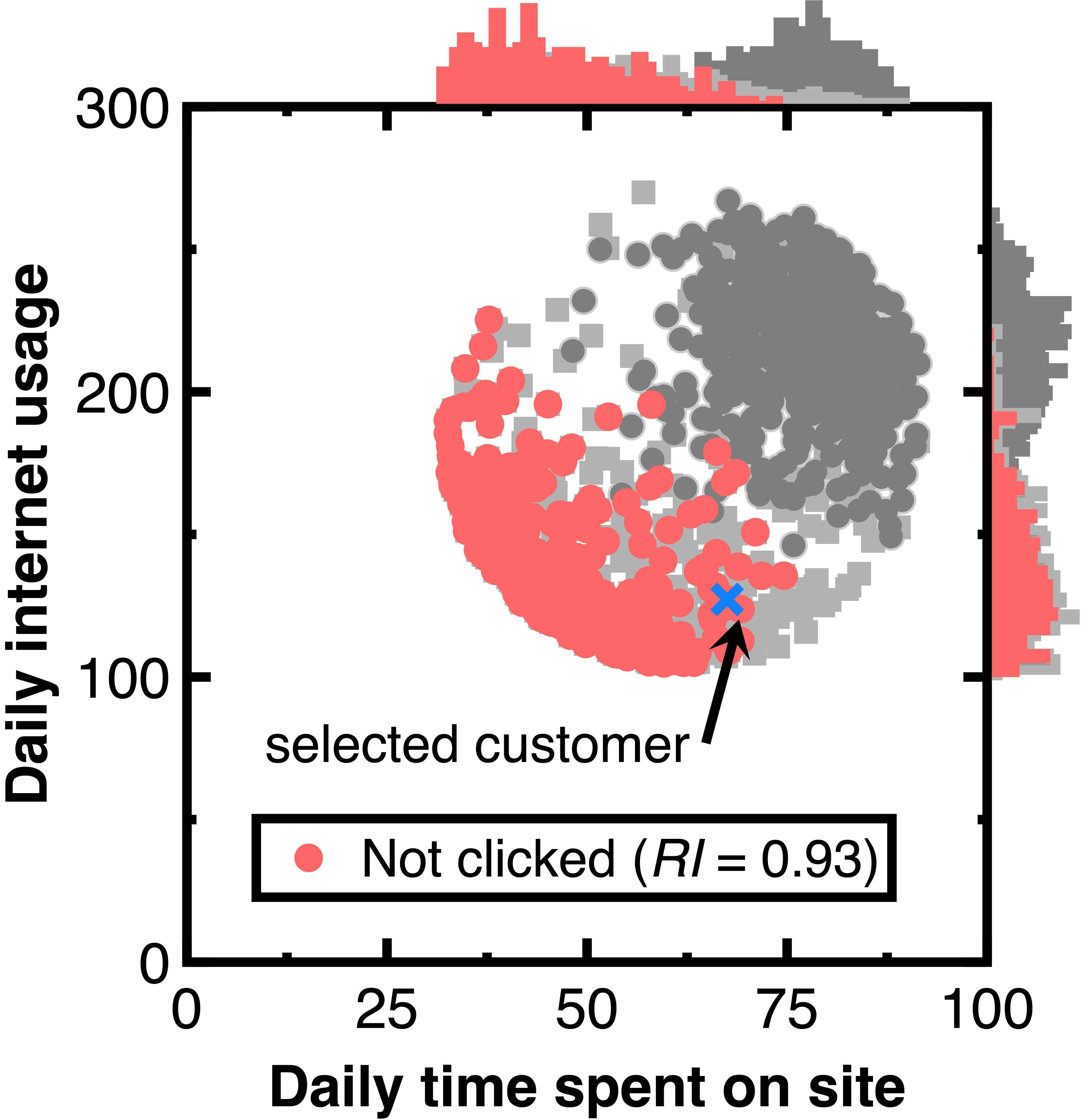}
\caption{Audit evidence relevant to a chosen customer ('Daily time spent on site' = 67.51, 'Age' = 43, 'Area in-come' = 23942.61, 'Daily internet usage' = 127.2, and 'Clicked on Ad' = 'Not clicked') }
\end{figure} 
\subsection{Experiment 2}

A spam message is one of the unstructured data that did not appear in the conventional sampling. In this experiment, this study introduces a data set containing 5572 messages, and 13 \% of them are spam. This study randomly selects 75 \% of them as train data. The other 25 \% are test data. In implementing this experiment, the first step is preprocessing these train and test data by vectorizing each message into a series of keywords. We employ a dictionary to select candidate keywords. Counting their frequencies is next performed. Classifying ham and spam messages is done by setting a class variable $C_i\;(1 \leq i \leq N)$ indicating a spam or ham message, and attributes are the frequency of keywords.

Based on the counts of keywords in ham and spam messages of experimental data, Figure 7 compares the top 20 keywords. Choosing them eliminates ordinary conjunctions and prepositions such as 'to' and 'and.' We can understand the unique keywords of spam messages from Figure 7.

To prevent sampling frame errors and undercoverage [15], Figure 8 compares the corresponding ROC curves versus different machine learning algorithms. It also shows the confusion matrix output by Equations (3)-(4). We have normalized its components based on the amount of test data. Table 2 lists other metrics for demonstrating classification accuracy on this confusion matrix.

Calculating the AUC values from Figure 8 yields 0.989 (Equations (3)-(4)), 0.923 (Random forest classifier), and 0.934 (Support vector machines model with a radial basis function kernel). Such AUC values indicate a support vector machines model, random forest, and Equations (3)-(4) are all good models for preventing sampling frame errors and undercoverage; however, the performance of Equations (3)-(4) is still the best.

Next, this study chooses the 75 \% confidence interval of spam messages to generate audit evidence. We obtained 652 samples of spam messages. Figure 9 compares counts of the top 20 keywords of original text data and audit evidence. Substituting their posterior probabilities to compute the representativeness index $RI$ equals 0.997. 

Figure 9 demonstrates that machine learning integration promotes sampling unstructured data (e.g., spam messages) while keeping their crucial information. The design of conventional sampling methods does not consider unstructured data [4]. In this figure, sampling spam messages keeps the ranking of all the top 20 keywords. The resulting samples may form a benchmark data set for testing the performance of different spam message detection methods.
\begin{figure}[H]
	\begin{adjustwidth}{-\extralength}{0cm}
		\centering
		\includegraphics[width=14cm]{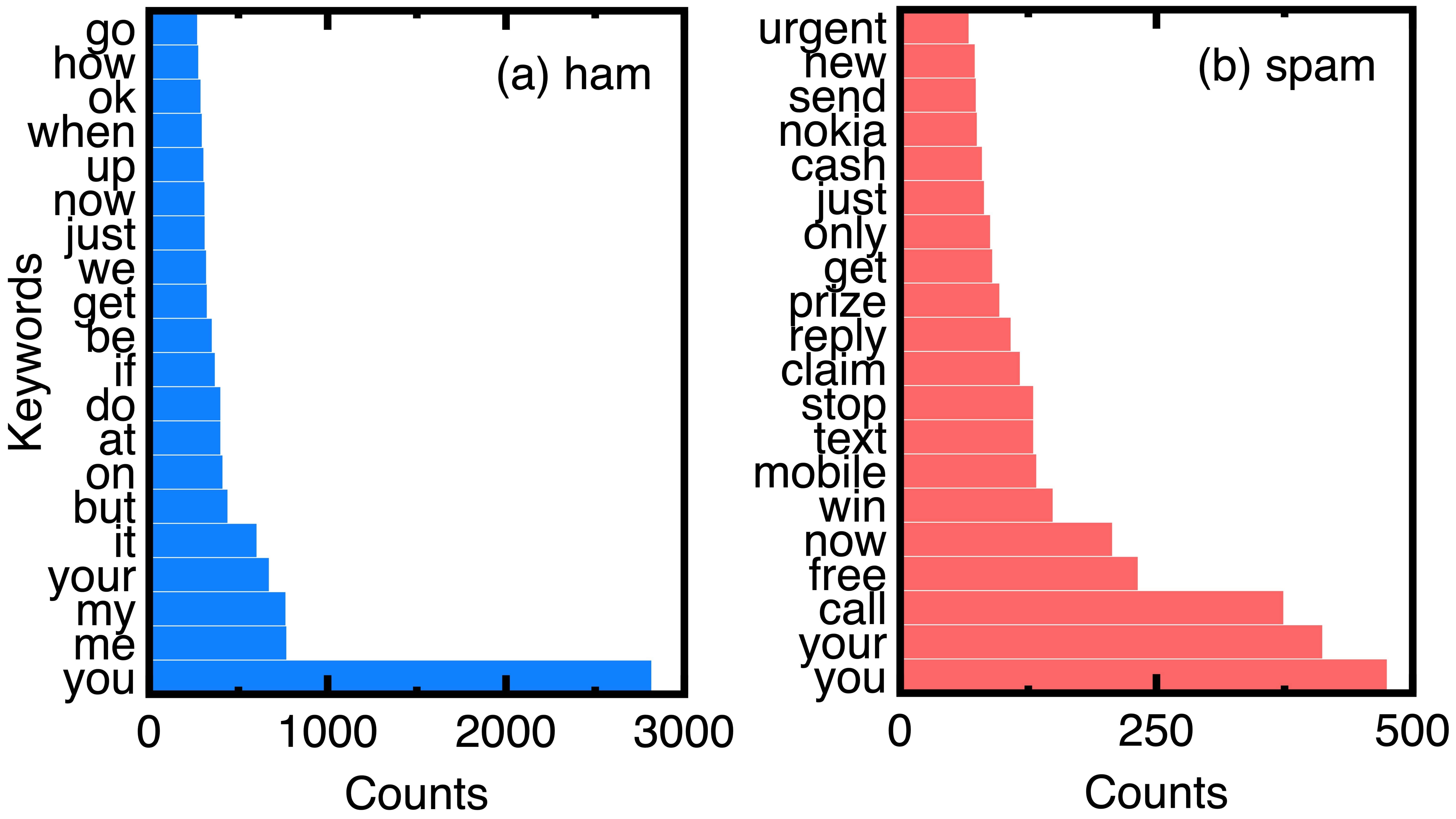}
	\end{adjustwidth}
	\caption{Comparison of top 20 keywords in ham and spam messages: (a) ham messages; (b) spam messages}
\end{figure} 
\begin{figure}[H]
	\includegraphics[width=8 cm]{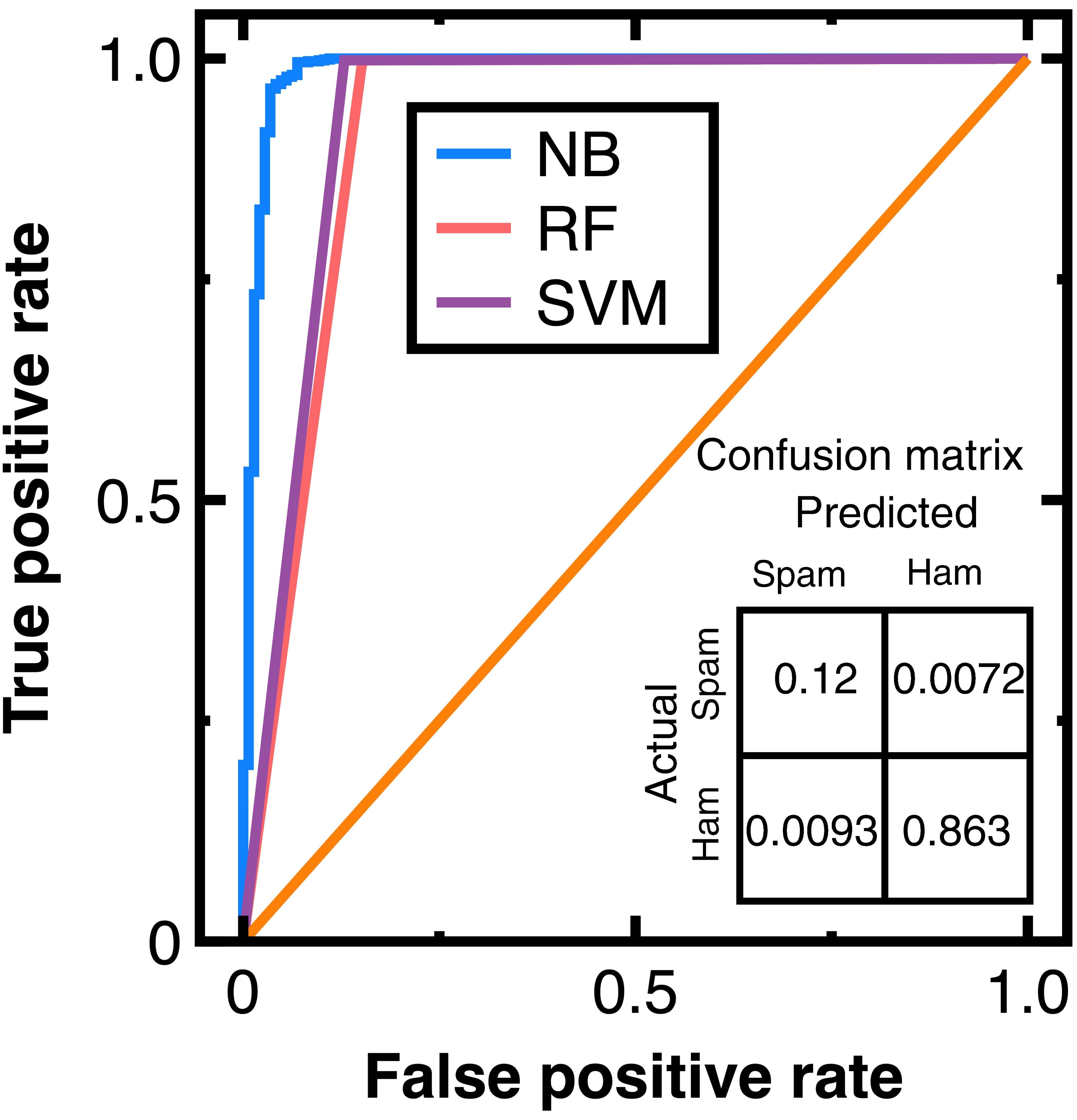}
	\caption{ROC curves provided by different machine learning algorithms and the confusion matrix output by Equation (3)-(4) for Experiment 2}
\end{figure}
\begin{table}[H] 
	\caption{Metrics output by Equations (3)-(4) for Experiment 2}
	\newcolumntype{C}{>{\centering\arraybackslash}X}
	\begin{tabularx}{\textwidth}{CC}
		\toprule \textbf{Metric}
		& \textbf{Value}\\
		\midrule
		Accuracy & 0.983\\
		Precision & 0.992\\
		Recall & 0.989\\
		Specificity & 0.992 \\
		F1 score & 0.99\\
		\bottomrule
	\end{tabularx}
\end{table}
\begin{figure}[H]
	\includegraphics[width=8 cm]{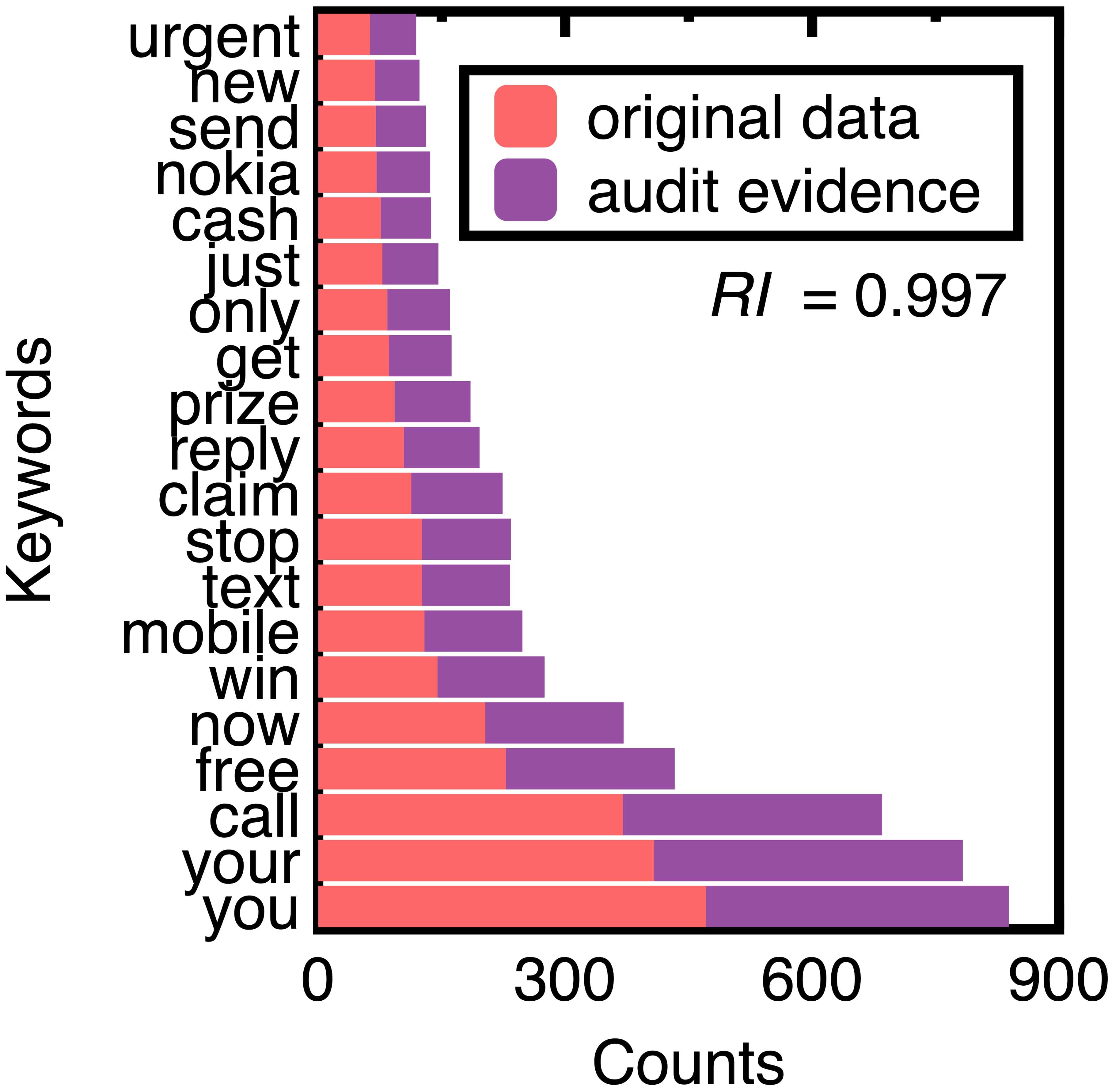}
	\caption{Comparison of top 20 keywords in original text data and audit evidence}
\end{figure}
\subsection{Experiment 3}
The third experiment illustrates that integrating machine learning with sampling can balance representativeness and riskiness. We use the Panama Papers to create a directed graph model having 535891 vertices in which each vertex denotes a suspicious financial account. Its attributes are the degree centrality and clustering coefficient.

The Panama Papers were a massive leak of documents. They exposed how wealthy individuals, politicians, and public figures worldwide used offshore financial accounts and shell companies to evade taxes, launder money, and engage in other illegal activities.

The degree centrality $D$ [17] is the number of edges connecting to a vertex. The higher the degree centrality, the greater the possibility detects black money flows. Besides, we consider that two financial accounts may have repeated money transfers. Therefore, computing the degree centrality considers the existence of multiple edges. For example, if a sender transfers money to a payee two times, the degree of a vertex simulating such a sender or payee equals 2.

Meanwhile, the clustering coefficient $c$ [17] measures the degree to which nodes in a graph tend to group. Evidence shows that in real-world networks, vertices may create close groups characterized by a relatively high density of ties. In a money laundering problem, a unique clustering coefficient may highlight a group within which its members exchange black money. Like the computation of degree centrality, calculating the clustering coefficient considers the possible existence of multiple edges.

The purpose of generating Experiment 3 is to demonstrate that integrating machine learning with sampling can balance representativeness and riskiness. Therefore, we set the $C_i\;(1\leq i \leq N)$ variable according to the $D_i$ and $c_i$ values. Table 3 lists the results. Its final column lists the total members corresponding to each $C_i$ class.
\begin{table}[H] 
	\caption{The resulting degree centrality $D_i$, clustering coefficient $c_i\;(1 \leq i \leq N)$, and total number of members in each $C_i$ class}
	\newcolumntype{C}{>{\centering\arraybackslash}X}
	\begin{tabularx}{\textwidth}{CCCC}
		\toprule \textbf{Class variable $c_i$}
		& \textbf{Degree centrality $D_i$ }&\textbf{Clustering coefficient  $c_i$} & \textbf{Total number of members}\\
		\midrule
		$1$ & [0,2)&[0,1]&338800\\
		$2$  &[2,4)&[0,1]&117323\\
		$3$  &[4,6)&[0,0.417]&41720\\
		$4$  &[6,10)&[0,0.367]&22743\\
		$5$  &[10,$\infty$)&[0,0.28]&15304\\
		\bottomrule
	\end{tabularx}
\end{table}

To prevent sampling frame errors and undercoverage [15], Figure 10 compares the ROC curves output by different machine learning algorithms in classifying nodes in Experiment 3. Obtaining Figure 10 chooses 80 \% of random nodes as train data and other vertices as test data. Moreover, Equations (3)-(4) output the confusion matrix shown in Equation (18):
\begin{equation}
\left[{\begin{array}{ccccc}{0.6311}&{0}&{0}&{0}&{0}\\{0}&{0.2198}&{0}&{0}&{0}\\{0}&{0.00139}&{0.077}&{0}&{0}\\{0}&{0}&{0.00031}&{0.0042}&{0}\\{0}&{0}&{0}&{0.004}&{0.0244}\end{array}}\right]
\end{equation}
in which each component has been normalized based on the amount of test data. 

From Equation (18), we further calculate the averaged accuracy, specificity, recall, precision, and F1 value, as shown in Table 4. Next, calculating the AUC values from Figure 10 and Table 4 results in 0.965 (Equations (3)-(4)), 0.844 (Random forest classifier), and 0.866 (Support vector machines model with a radial basis function kernel). Figure 10 indicates that the random forest classifier and support vector machines model with a radial basis function kernel are unsuitable for this experiment. Since we have a high volume of data in this experiment, these two algorithms may output unacceptable errors in sampling nodes.
\begin{figure}[H]
	\includegraphics[width=8 cm]{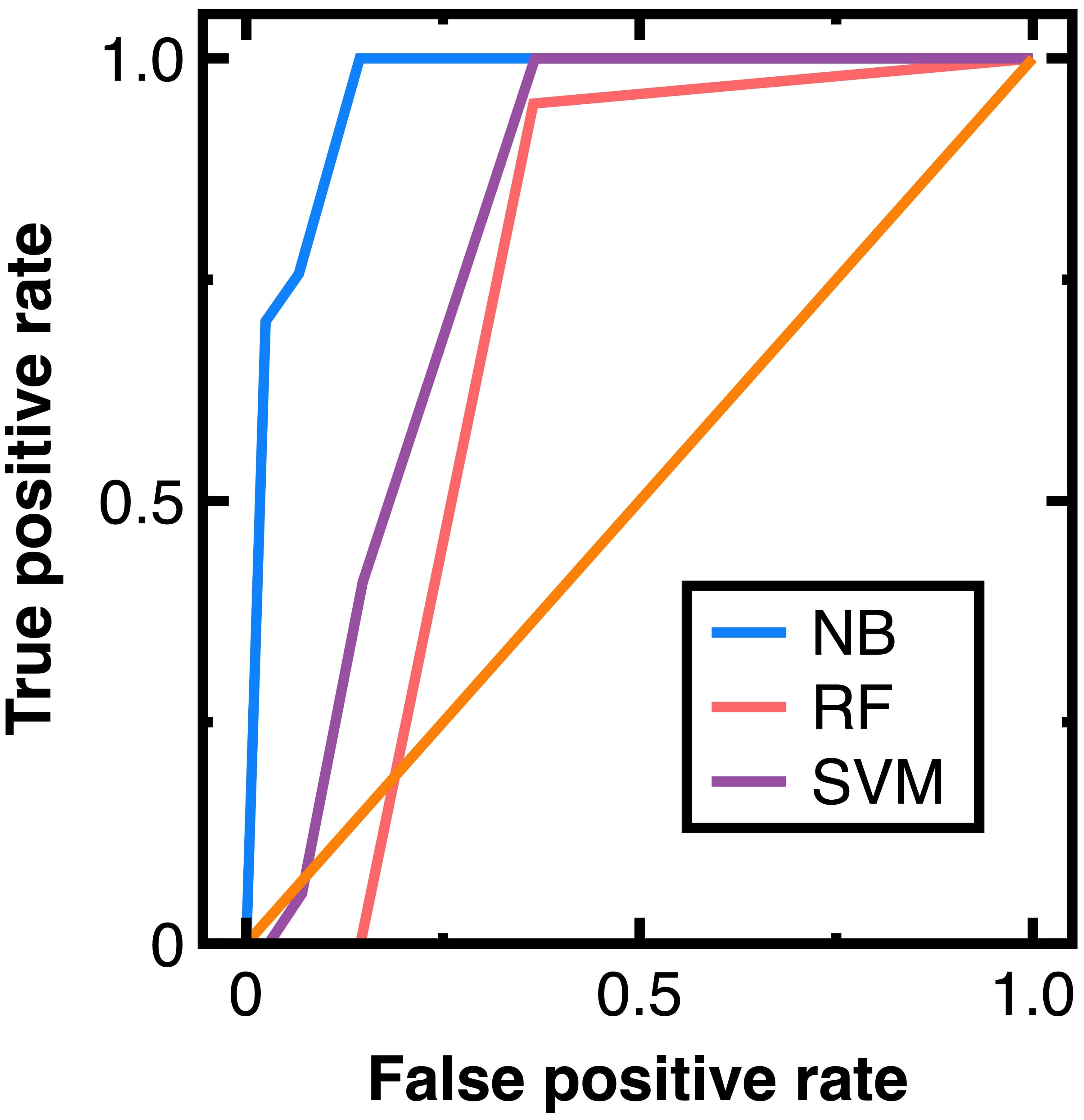}
	\caption{ROC curves provided by different machine learning algorithms for Experiment 3}
\end{figure}
\begin{table}[H] 
	\caption{Metrics calculated from Equation (18)}
	\newcolumntype{C}{>{\centering\arraybackslash}X}
	\begin{tabularx}{\textwidth}{CC}
		\toprule \textbf{Metric}
		& \textbf{Averaged value}\\
		\midrule
		Accuracy & 0.995\\
		Precision & 0.992\\
		Recall & 0.989\\
		Specificity & 0.992 \\
		F1 score & 0.99\\
		\bottomrule
	\end{tabularx}
\end{table}

Suppose a 75 \% confidence interval to sample members of each class $C_i\;(i = 1, 2\ldots,N)$. However, we agree that the $C_i = 5$ class has the risker members. High $D_i$ values imply frequent transactions. Therefore, further drawing audit evidence from samples with $\Pr\left({{{C}_{i}{=}{5}}\vert{{X}_{j}}}\right){(}{1}\leq{i}{,}{j}\leq{N}{)}=1$ values within the 75 \% confidence interval of the $C_i = 5$ class. The red points in Figure 11 represent the resulting audit evidence. Heavy gray points denote original data. The legend of this figure lists the corresponding representativeness index $RI$ and the number of drawn samples.

Carefully inspecting Figure 11 indicates that vertices ($D_i \geq 13\;(i = 1,2\ldots,N)$) are drawn as audit evidence. They are riskier than other nodes in the $C_i = 5$ class. With the help of a Naive Bayes classifier (Equations (3)-(4)), profiling the class $C_i = 5$ is unnecessary before sampling this $C_i = 5$ class. This unnecessity illustrates the difference between sampling with machine learning integration and conventional sampling methods. 
\begin{figure}[H]
	\includegraphics[width=8 cm]{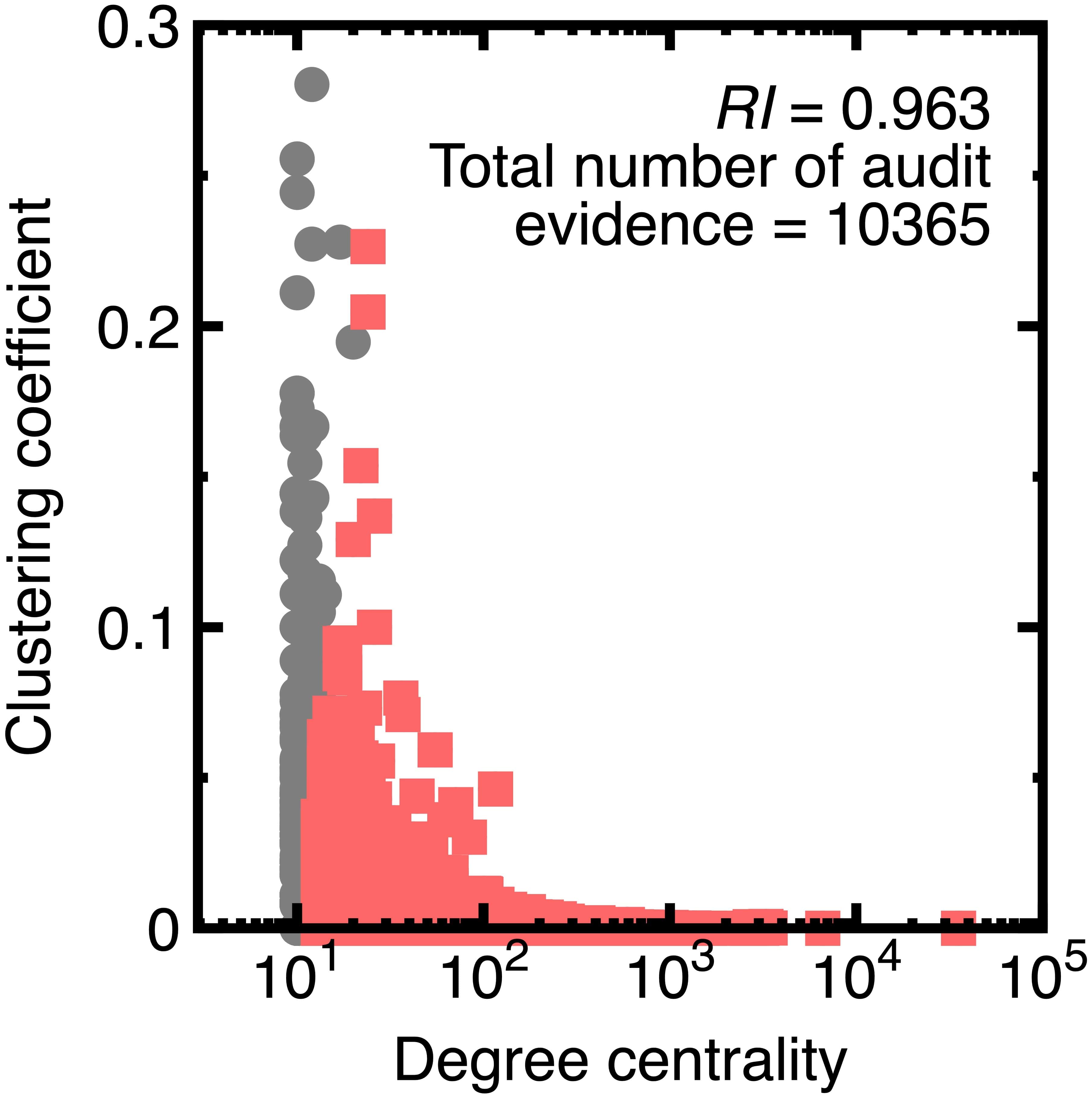}
	\caption{Risker audit evidence for Experiment 3}
\end{figure}
\section{Discussion}

Section 4 implies the benefits and limitations of integrating a Naive Bayes classifier with sampling. We further list these benefits and limitations:
\begin{itemize}
  \item Conventional sampling methods [4] may not profile the full diversity of data; thus, they may provide biased samples. Since this study samples data after classifying them using a Naive Bayes classifier, it substitutes for a sampling method to profile the whole diversity of data. Experimental results of Section 4 indicate that the Naive Bayes classifier classifies three open data sets accurately, even if they are excessive. Those accurate classification results indicate that we capture the whole diversity of experimental data.
  \item Developing conventional sampling methods may not consider complex patterns or correlations in data [4]. In this study, we handle complex correlations or patterns in data (for example, a graph structure in Section 4.3) by a Naive Bayes classifier. This design mitigates the sampling bias caused by complex patterns or correlations if it provides accurate classification results.
  \item Section 4.3 indicates that a Naive Bayes classifier works well for big data in a money laundering problem. It outperforms the random forest classifier and support vector machines model with a radial basis function kernel in classifying massive vertices. Thus, we illustrate that the efficiency of sampling big data can be improved. One can sample risker nodes modeling fraudulent financial accounts without profiling specific groups of nodes.
  \item Development of conventional sampling methods considers structured data; however, they struggled to handle unstructured data such as spam messages in Section 4.2. We resolve this difficulty by employing a Naive Bayes classifier before sampling.
  \item Since this study samples data from each class classified by a Naive Bayes classifier, accurate classification results eliminate sample frame errors and improper sampling sizes. 
  \end{itemize}
Nevertheless, this study also finds limitations in integrating machine learning and sampling. They are listed as follows:
\begin{itemize}
  \item It is still possible that a Naive Bayes classifier provides inaccurate classification results. One should test the classification accuracy before sampling with machine learning integration.
  \item In implementing Section 3.2, thresholds $\sigma_j\;(j = 1-3)$ are needed. However, we should inspect variations of the prior probabilities for determining proper $\sigma_j\;(j = 1-3)$ values. They denote the second limitation of our machine learning-based sampling. \end{itemize}
\section{Conclusions}

Sampling plays a crucial role in auditing. It provides a mechanism for auditors to draw audit evidence. However, various challenges exist within available sampling methodologies, including selection bias, sampling frame errors, improper sampling sizes, and handling of unstructured and massive data. This study develops a Naive Bayes classifier as a sampling tool for drawing data within a confidence interval symmetric to the median or sampling asymmetrically riskier samples. It is employed to overcome the challenges mentioned above. From Section 4, we conclude that sampling with machine learning integration has the benefits of providing unbiased samples, handling complex patterns or correlations in data, processing unstructured or big data, and avoiding sampling frame errors or improper sampling sizes.

However, sampling using a Naive Bayes classifier has limitations. Inaccurate classification results output by the Naive Bayes classifier may result in biased samples or sampling frame errors. Overcoming them requires testing the Naive Bayes classifier before applying it to sampling.

\vspace{6pt} 




\authorcontributions{Conceptualization, G.Y. Sheu; methodology, G. Y. Sheu; software, G. Y. Sheu; validation, N. R. Liu; formal analysis, G. Y. Sheu; investigation, G. Y. Sheu; resources, G. Y. Sheu; data curation, N. R. Liu; writing---original draft preparation, N. R. Liu; writing---review and editing, G. Y. Sheu; visualization, N. R. Liu; supervision, G. Y. Sheu; project administration, G. Y. Sheu; funding acquisition, N. R. Liu}

\funding{The implementation of this article is funded in part by the National Science and Technology Council, R.O.C., under Grant No. 112-2813-C-309-002-H.}

\dataavailability{Customer ad click prediction dataset at \url{https://www.kaggle.com/code/mafrojaakter/customer-ad-click-prediction}; SMS   spam collection dataset at \url{https://www.kaggle.com/code/mafrojaakter/customer-ad-click-prediction}; Panama Papers at \url{https://offshoreleaks.icij.org/pages/database}.} 



\conflictsofinterest{The authors declare no conflicts of interest.} 

\reftitle{References}

\PublishersNote{}

\begin{thebibliography}{999}
\bibitem[Author2(year)]{ref-book1}
Deng, H.; Sun Y.; Chang, Y.; Han, J. \textit{Probabilistic Models for Classification}, In {\em Data Classification: Algorithms and Applications}; Aggarwal, C.C. Eds.; Chapman and Hall/CRC: New York, USA, 2014; pp. 65--86.
\bibitem[Author1(year)]{ref-journal1}
Schreyer, M.; Gierbl, A.; Ruud, T.F.; Borth, D. Artificial intelligence enabled audit sampling -- Learning to draw representative and interpretable audit samples from large-scale journal entry data. {\em Expert Focus} {\bf 2022}, {\em 04}, 106--112.
\bibitem[Author1(year)]{ref-journal2}
Zhang, Y.; Trubey, P. Machine learning and sampling scheme: An empirical study of money laundering detection. {\em Comput. Econ.} {\bf 2019}, {\em 54(3)}, 1043--1063.
\bibitem[Author1(year)]{ref-journal3}
Aitkazinov, A. The role of artificial intelligence in auditing: Opportunities and challenges. {\em Int. J. Res. Eng. Sci. Manag.} {\bf 2023}, {\em 6(6)}, 117--119.
\bibitem[Author1(year)]{ref-journal4}
Chen, Y.; Wu, Z.; Yan, H. A full population auditing method based on machine learning. {\em Sustainability} {\bf 2022}, {\em 14(24)}, 17008.
\bibitem[Author1(year)]{ref-journal5}
Bertino, S. A measure of representativeness of a sample for inferential purposes. {\em Int. Stat. Rev.} {\bf 2006}, {\em 74(2)}, 149--159.
\bibitem[Author3(year)]{ref-book2}
Guy, D.M.; Carmichael, D.R.; Whittington, O.R. \textit{Audit Sampling: An Introduction to Statistical Sampling in Auditing}, 5th ed.; John Wiley \& Sons: New York, USA, 2001.
\bibitem[Author6(year)]{ref-proceeding1}
Schreyer, M.; Sattarov, T.; Borth, D. Multi-view contrastive self-supervised learning of accounting data representations for downstream audit tasks. In Proceedings of the Second ACM International Conference on AI in Finance Virtual Event, New York, USA, 5 3 2021; DOI: 10.1145/3490354.3494373.
\bibitem[Author1(year)]{ref-journal6}
Schreyer, M.; Sattarov, T.; Reimer, G.B.; Borth, D. Learning sampling in financial statement audits using vector quantised autoencoder. {\em arXiv} {\bf 2020}, DOI: 10.48550/arXiv.2008.02528.
\bibitem[Author1(year)]{ref-journal7}
Lee, C. Deep learning-based detection of tax frauds: an application to property acquisition tax. {\em Data Technol. Appl.} {\bf 2022}, {\em 56(3)}, 329--341.
\bibitem[Author1(year)]{ref-journal8}
Chen, Z.; Li, C.; Sun, W. Bitcoin price prediction using machine learning: An approach to sample dimensional engineering. {\em J. Comput. Appl. Math.} {\bf 2020}, {\em 365}, 112395.
\bibitem[Author6(year)]{ref-proceeding1}
Liberty, E.; Lang, K.; Shmakov, K. Stratified sampling meets machine learning. In Proceedings of the 33rd International Conference on Machine Learning, New York, USA, 6 19 2016.
\bibitem[Author1(year)]{ref-journal9}
Hollingsworth, J.; Ratz, P.; Tanedo, P.; Whiteson, D. Efficient sampling of constrained high-dimensional theoretical spaces with machine learning. {\em Eur. Phys. J. C} {\bf 2021}, {\em 81(12)}, 1138.
\bibitem[Author1(year)]{ref-journal10}
Artrith, N.; Urban, A.; Ceder, G. Constructing first-principles diagrams of amorphous Li\textsubscript{x}Si using machine-learning-assisted sampling with an evolutionary algorithm. {\em J. Chem. Phys.} {\bf 2018}, {\em 148(24)}, 241711.
\bibitem[Author1(year)]{ref-journal11}
Huang, F.; No, W.G.; Vasarhelyi, M.A.; Yan, Z. Audit data analytics, machine learning, and full population testing. {\em J. Finance Data Sci.} {\bf 2022}, {\em 8}, 138--144.
\bibitem[Author1(year)]{ref-journal12}
Kolmogorov, A. Sulla determination empirica di una legge di distribuzione. {\em G. Inst. Ital. Attuari.} {\bf 1933}, {\em 4}, 83--91.
\bibitem[Author3(year)]{ref-book3}
Wasserman, S.; Faust, K. \textit{Social Network Analysis: Methods and Applications}, 1st ed.; Cambridge University Press: Cambridge, New York, USA, 1994.
\end{thebibliography}
\end{document}